\def\year{2020}\relax
\documentclass[letterpaper]{article} 
\usepackage{aaai20}  
\usepackage{times}  
\usepackage{helvet} 
\usepackage{courier}  
\usepackage[hyphens]{url}  
\usepackage{graphicx} 
\usepackage{graphicx}  
\usepackage{amsthm}
\usepackage{amsmath}
\usepackage{amsfonts}       
\usepackage{nicefrac}       
\usepackage{microtype}      
\usepackage{amssymb }
\usepackage{booktabs}
\usepackage{algorithmic}
\usepackage{algorithm}
\usepackage{tcolorbox}
\newcommand\numberthis{\addtocounter{equation}{1}\tag{\theequation}}
\renewcommand{\vec}[1]{\mathbf{#1}}
\usepackage{multirow}
\numberwithin{equation}{section}

\title{Multi-Fidelity Multi-Objective Bayesian Optimization:  \\ An Output Space Entropy Search Approach}
\author{Syrine Belakaria, Aryan Deshwal, Janardhan Rao Doppa\\ 
School of EECS, Washington State University\\ 
\text{\{syrine.belakaria, aryan.deshwal, jana.doppa\}@wsu.edu}}
\begin{document}
\maketitle
\begin{abstract}

We study the novel problem of blackbox optimization of multiple objectives via multi-fidelity function evaluations that vary in the amount of resources consumed and their accuracy. The overall goal is to approximate the true Pareto set of solutions by minimizing the resources consumed for function evaluations. For example, in power system design optimization, we need to find designs that trade-off cost, size, efficiency, and thermal tolerance using multi-fidelity simulators for design evaluations. In this paper, we propose a novel approach referred as {\em {\bf M}ulti-{\bf F}idelity {\bf O}utput {\bf S}pace {\bf E}ntropy Search for {\bf M}ulti-objective {\bf O}ptimization (MF-OSEMO)} to solve this problem. The key idea is to select the sequence of candidate input and fidelity-vector pairs that maximize the information gained about the true Pareto front per unit resource cost. 
Our experiments on several synthetic and real-world  benchmark  problems  show  that MF-OSEMO, with both approximations, significantly improves over the state-of-the-art single-fidelity algorithms for multi-objective optimization.

\end{abstract}

\section{Introduction}

Multi-objective optimization of expensive black-box functions has many real-world applications. For example, creating hardware to optimize performance, reliability, and thermal objectives. There are two key challenges in solving these problems. First, the objective functions are unknown and we need to select from experiments of different fidelity to evaluate each candidate input. These multi-fidelity experiments vary in the amount of resources consumed and the accuracy of evaluation.
Second, all the objectives cannot be optimized simultaneously due to their conflicting nature. Hence, we resort to finding the {\em Pareto optimal} set of solutions. A solution is called Pareto optimal if it cannot be improved in any of the objectives without compromising some other objective. 
The overall goal is to approximate the optimal Pareto set by minimizing the overall resource cost of function evaluations.

Bayesian optimization (BO) \cite{BO-Survey} is a popular framework for solving blackbox optimization problems. BO methods build a surrogate statistical model, e.g., Gaussian process, from the training data of function evaluations; employ a acquisition function (AF) that is parameterized by the model, e.g., upper-confidence bound, to score the utility of evaluating candidate inputs; and select the highest scoring input for evaluation in each iteration. Existing AFs can be broadly classified into two categories. First, {\em myopic} AFs rely on improving a ``local'' measure of utility (e.g., expected improvement). Second, {\em non-myopic} AFs measure the ``global'' utility of evaluating a candidate input for solving the black-box optimization problem (e.g., predictive entropy search). Prior work has shown the advantages of non-myopic AFs over myopic AFs in terms of both theory and practice \cite{jiang2017efficient,PES,MES,hoffman2015output}.

In this paper, we propose a novel and principled approach referred as {\bf M}ulti-{\bf F}idelity {\bf O}utput {\bf S}pace {\bf E}ntropy Search for {\bf M}ulti-objective {\bf O}ptimization (MF-OSEMO) to solve multi-objective optimization problems via multi-fidelity function evaluations. {\em To the best of our knowledge, this is the first work to study this problem within ML literature}. MF-OSEMO employs an output space entropy based non-myopic acquisition function to select the candidate inputs and fidelity vectors for evaluation. Output space entropy search has many advantages over other non-myopic AFs based on input space entropy \cite{hoffman2015output,MES}: a) allows much tighter approximation; b) significantly cheaper to compute; and c) naturally lends itself to robust optimization. We provide two qualitatively different approximations to efficiently compute the entropy, which is a key step for MF-OSEMO. These approximations make different trade-offs in terms of accuracy and computational-efficiency: one has a closed-form expression and another employs numerical integration.
\vspace{1.0ex}

\noindent {\bf Contributions.} We make the following key contributions.

\begin{itemize}
\setlength\itemsep{0em} 
\item Developing a principled approach referred as MF-OSEMO to solve multi-fidelity multi-objective blackbox optimization problems. MF-OSEMO employs an output space entropy based acquisition function to select the sequence of candidate inputs and fidelity vectors for evaluation. Providing two different approximations within MF-OSEMO.
\item Experimental evaluation on synthetic and real-world benchmark problems to show the effectiveness of MF-OSEMO over state-of-the-art single-fidelity algorithms. 
\end{itemize}

\section{Problem Setup}

Let $\mathfrak{X} \subseteq \Re^d$ be an input space. In the multi-objective optimization problem, our goal is to minimize $K \geq 2$ {\em expensive} objective functions $f_1(\vec{x}),f_2(\vec{x}),\cdots,f_K(\vec{x})$. Evaluation of a candidate input $\vec{x} \in \mathfrak{X}$  produces a vector of $K$ function values $\vec{y} = (y_1, y_2,\cdots,y_K)$, where $y_i = f_i(x)$ for all $i \in \{1,2, \cdots, K\}$.  A point $\vec{x}$ is said to {\em Pareto-dominate} another point $\vec{x}^\prime$ if $f_i(\vec{x}) \leq f_i(\vec{x}^\prime)$  $\forall{i}$ and there exists some $j \in \{1, 2, \cdots,K\}$ such that $f_j(\vec{x}) < f_j(\vec{x}^\prime)$. The optimal solution of MOO problem is a set of points $\mathcal{X}^* \subset \mathfrak{X}$ such that no point $\vec{x}^\prime \in \mathfrak{X} \setminus \mathcal{X}^*$ Pareto-dominates a point $\vec{x} \in \mathcal{X}^*$. The solution set $\mathcal{X}^*$ is called the optimal {\em Pareto set} and the corresponding set of function values $\mathcal{Y}^*$ is called the optimal {\em Pareto front}. In the multi-fidelity version of MOO problem, we have access to $M_i$ fidelities for each function $f_i$ that vary in the amount of resources consumed and the accuracy of evaluation. Let $\lambda_i^{(m_i)}$ be the cost of evaluating $i^{th}$ function $f_i$ at $m_i \in [M_i]$ fidelity, where $m_i$=$M_i$ corresponds to the highest fidelity for $f_i$. Evaluation of an input $\vec{x}\in \mathfrak{X}$ with fidelity vector $\vec{m} = [m_1, m_2, \cdots, m_K]$ produces an evaluation vector of $K$ values denoted by $\vec{y}^{\vec{m}} \equiv [y_1^{(m_1)}, \cdots, y_K^{(m_K)}]$, where $y_i^{(m_i)} = f_i^{(m_i)}(x)$ for all $i \in \{1,2, \cdots, K\}$, and normalized cost of evaluation is $\lambda^{(\vec{m})} \equiv \sum_{i=1}^{K} \left({\lambda_i^{(m_i)}}/{\lambda_i^{(M_i)}}\right)$. We normalize the total cost since the cost units can be different for different objectives(e.g. cost unit for $f_1$ is computation time while cost unit for $f_2$ could be memory space size). 
Our goal is to approximate $\mathcal{X}^*$ by minimizing the overall cost of function evaluations. For the sake of reader, Table~\ref{table:notations} contains all the mathematical notations used in this paper.

\begin{table*}[t]
    \centering
    \resizebox{0.75\linewidth}{!}{
    \begin{tabular}{|c|c|}
    \hline
    {\bf Notation} & {\bf Meaning} \\
    \hline \hline
       $\vec{x}, \vec{y}, \vec{f}, \vec{m}$ & bold notation represents vectors\\
       \hline
       $[n]$ & set of first $n$ natural numbers $\{1,2,\cdots, n\}$ \\
       \hline
       $f_1, f_2, \cdots, f_K$  & true objective functions \\
       \hline
       $M_1, M_2, \cdots, M_K$ & no. of fidelities for each function \\
       \hline
       $\vec{x}$ & input vector \\
       \hline
    $\vec{m} = [m_1, m_2, \cdots, m_K]$& fidelity vector where each fidelity $m_j \in [M_j]$ \\
       \hline
        $y_j^{m_j}$ &$j$th function $f_j$ evaluated at $m_j$th fidelity  where $m_j \in [M_j]$\\
       \hline
      $\vec{y}^{\vec{m}}$   & output vector equivalent to $ [y_1^{(m_1)}, \cdots, y_K^{(m_K)}]$\\  
      \hline 
$\mathcal{Y}^*$ & true Pareto front of the highest fidelities $[y_1^{(M_1)}, y_2^{(M_2)}, \cdots, y_K^{(M_K)}]$\\      \hline 
$\lambda_j^{({m_j})}$ & cost of evaluating $j$th function $f_j$ at $m_j$th fidelity\\
\hline
$\lambda^{(\vec{m})}$ & total normalized cost $\lambda^{(\vec{m})} \equiv \sum_{j=1}^{K} \left( {\lambda_j^{(m_j)}}/{\lambda_j^{(M_j)}}\right)$  \\
\hline
$\Tilde{f}_j^{(m_j)}$ & function sampled from $j$th gaussian process model at $m_j$th fidelity\\
\hline
    \end{tabular}}
    \caption{Table describing the mathematical notations used in this paper.}
    \label{table:notations}
\end{table*}

\section{Related work}
\begin{figure}
    \centering
    \includegraphics[width=0.7\columnwidth]{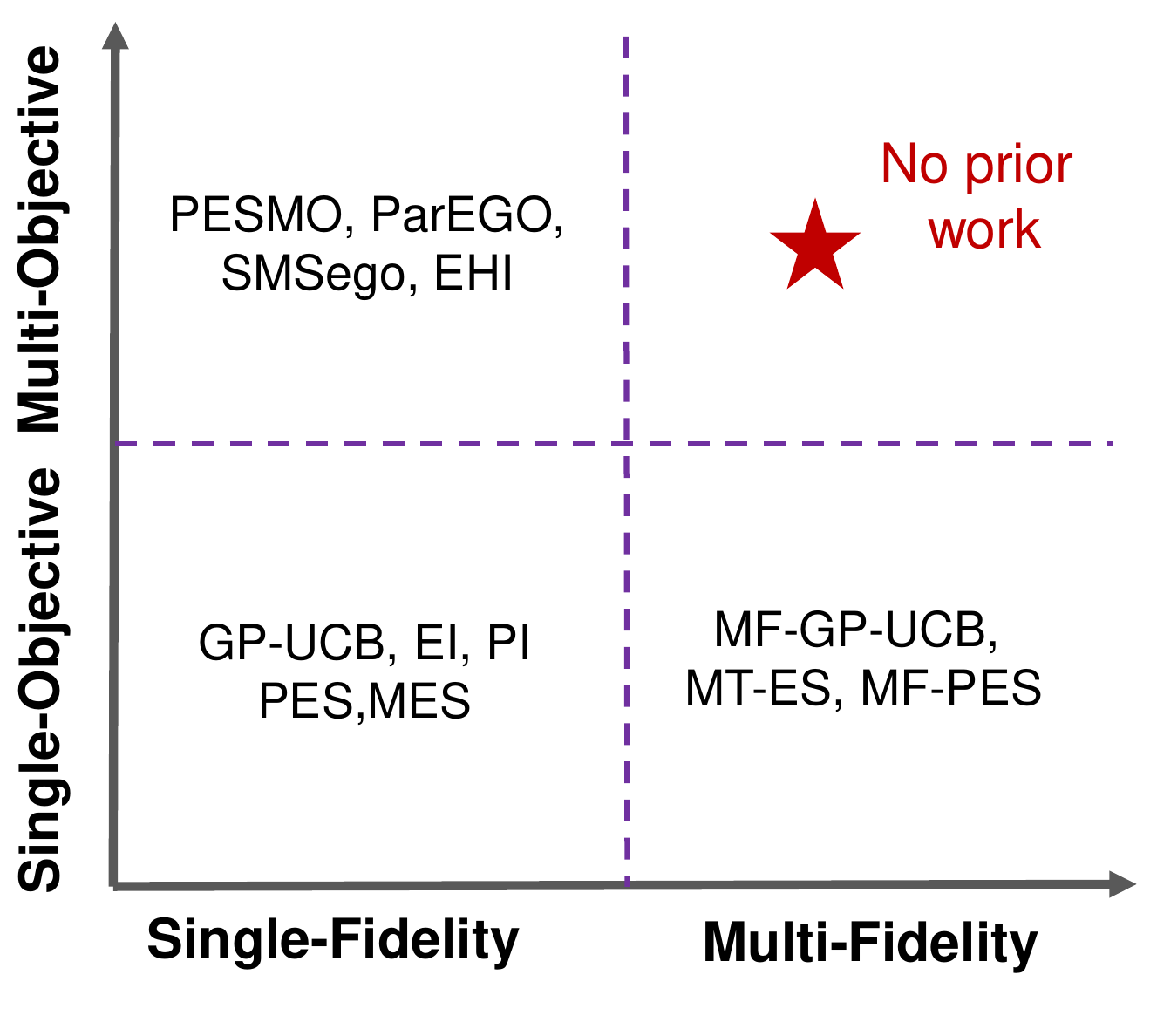}
    \caption{Current state of knowledge for generic BO methods.}
    \label{fig:relatedwork}
\end{figure}
\vspace{1.0ex}

\noindent {\bf Multi-fidelity single-objective optimization.}
AFs for single-fidelity and single-objective BO has been extensively studied \cite{BO-Survey}. 
Canonical examples of myopic AFs include expected improvement (EI) and upper-confidence bound (UCB). EI was extended to multi-fidelity setting \cite{huang2006sequential,picheny2013quantile,lam2015multifidelity}. The popular GP-UCB method \cite{gp-ucb} was also extended to multi-fidelity setting with discrete fidelities \cite{kandasamy2016gaussian} and continuous fidelities \cite{kandasamy2017multi}. 

Entropy based methods fall under the category of {\em non-myopic} AFs 
Some examples include entropy search (ES) \cite{entropy_search} and predictive entropy search (PES) \cite{PES}. Their multi-fidelity extensions include MT-ES \cite{swersky2013multi,klein2017fast} and MF-PES\cite{zhang2017information,mcleod2017practical}. Unfortunately, they inherit the computational difficulties of the original ES and PES. Max-value entropy search (MES) and output space predictive entropy search \cite{MES,hoffman2015output} are recent approaches that rely on the principle of output space entropy (OSE) search. Prior work \cite{MES} has shown advantages of OSE search in terms of compute-time, robustness, and accuracy over input space entropy search methods.
A recent work \cite{takeno2019multi} extended MES to multi-fidelity setting and showed its effectiveness over MF-PES. Recent work \cite{song2018general} proposed a general approach based on mutual information.

\vspace{1.0ex}


\noindent {\bf Single-fidelity multi-objective optimization.} Multi-objective algorithms can be classified into three families. {\em Scalarization methods} are model-based algorithms that reduce the problem to single-objective optimization. ParEGO method \cite{knowles2006parego} employs random scalarization for this purpose. 
ParEGO is simple and fast, but more advanced approaches often outperform it. {\em Pareto hypervolume optimization methods} optimize the Pareto hypervolume (PHV) metric \cite{emmerich2008computation} that captures the quality of a candidate Pareto set. This is done by extending the standard acquisition functions to PHV objective, e.g., expected improvement in PHV \cite{emmerich2008computation} and probability of improvement in PHV \cite{picheny2015multiobjective}. Unfortunately, algorithms to optimize PHV based acquisition functions scale very poorly and are not feasible for more than three objectives. To improve scalability, methods to reduce the search space are also explored  \cite{ponweiser2008multiobjective}. A common drawback of this family is that reduction to single-objective optimization 
can potentially lead to more exploitative behavior.

{\em Uncertainty reduction methods} like PAL \cite{zuluaga2013active}, PESMO \cite{PESMO} and the concurrent works USeMO \cite{Usemo} and MESMO \cite{MESMO} are principled algorithms based on information theory. 
In each iteration, PAL selects the candidate input for evaluation towards the goal of minimizing the size of uncertain set. PAL provides theoretical guarantees, but it is only applicable for input space $\mathfrak{X}$ with finite set of discrete points. USeMO is a general framework that iteratively generates a cheap Pareto front using the surrogate models and then selects the point with highest uncertainty as the next query.
PESMO relies on input space entropy
search and iteratively selects the input that maximizes the information gained about the optimal Pareto set $\mathcal{X}^*$. Unfortunately, optimizing this acquisition function poses significant challenges: a) requires a series of approximations, which can be potentially sub-optimal; and b) optimization, even after approximations, is expensive c) performance is strongly dependent on the number of Monte-Carlo samples. MESMO \cite{MESMO} is a concurrent work that improves over PESMO by extending MES to the multi-objective setting.

\vspace{1.0ex}

\noindent {\bf Application-specific multi-fidelity multi-objective optimization.} Prior work outside ML literature has considered domain-specific methods that employ single-fidelity multi-objective approaches in the context of multi-fidelity setting by using the lower fidelities {\em only as an initialization} \cite{kontogiannis2018comparison,ariyarit2017multi}. Specifically, \cite{ariyarit2017multi} employs the single-fidelity algorithm based on expected hypervolume improvement acquisition function and \cite{kontogiannis2018comparison} employs an algorithm that is very similar to SMSego. Additionally, both these methods model all fidelities with the same GP and assume that higher fidelity evaluation is a sum of lower-fidelity evaluation and offset error. These are strong assumptions and may not hold in general multi-fidelity settings including the problems we considered in our experimental evaluation. 
\section{MF-OSEMO Algorithm}
In this section, we explain the technical details of our proposed MF-OSEMO algorithm. 

\subsection{Surrogate models} 
\label{surrogatesection}
Let $D = \{(\vec{x}_i, \vec{y}_i^{(\vec{m})})\}_{i=1}^{t-1}$ be the training data from past $t{-1}$ function evaluations, where  
$\vec{x}_i \in \mathfrak{X}$ is an input and $\vec{y}^{(\vec{m})}_i = [y_1^{(m_1)},y_2^{(m_2)},\cdots,y_K^{(m_K)}]$ is the output vector resulting from evaluating functions $f_1^{(m_1)}, f_2^{(m_2)},\cdots,f_K^{(m_k)}$ at $\vec{x}_i$. 

Gaussian processes (GPs) are known to be effective surrogate models in prior work on single and multi-objective BO \cite{gp-ucb,PESMO}. 
We learn $K$ surrogate models $\mathcal{GP}_1,\mathcal{GP}_2,\cdots,\mathcal{GP}_K$ from $\mathcal{D}$, where each $\mathcal{GP}_j$ corresponds to the $j$th function $f_j$. In our setting, each function has multiple fidelities. So one ideal property desired for the surrogate model of a single function is to take into account all the fidelities in a single model. Multi-fidelity GPs (MF-GP) are capable of modeling functions with multiple fidelities in a single model. Hence, each of our surrogate model $\mathcal{GP}_j$ is a multi-fidelity GP.

Specifically, we use the MF-GP model as proposed in \cite{kennedy2000predicting,takeno2019multi}. We describe the complete details of the MF-GP model below.  One key thing to note about MF-GP model is that the kernel function ($k((\vec{x_i},m_i),(\vec{x_j},m_j))$) is dependent on both the input and the fidelity. 
For a given input $\vec{x}$, the MF-GP model returns a {\em vector} (one for each fidelity) of predictive mean, a {\em vector} of predictive variance, and a matrix of predictive covariance.
The MF-GP model has two advantages: 1) All fidelities are integrated into one single GP; and 2) Difference among fidelities are adaptively estimated without any additional feature representation for fidelities. It should be noted that we employ an independent multi-fidelity GP for each function.

\vspace{1.0ex}

\noindent \textbf{Multi-fidelity Gaussian process model.} We describe full details of a MF-GP model for one objective function $f_j$ (without loss of generality) below:

Let $y_j^{(1)}(\vec{x}), \ldots, y_j^{(M_j)}(\vec{x})$ represent the values obtained by evaluating the function $f_j$  at its $1$st, $2$nd, $\ldots, M_j$th  fidelity respectively.

In a MF-GP model, each fidelity is represented by a gaussian process and the observation is modeled as
\begin{align*}
 y_j^{(m_j)}(\vec{x}) = f_j^{(m_j)}(\vec{x}) + \epsilon, \quad \epsilon \sim \mathcal{N}(0, \sigma_{\rm noise}^2).
\end{align*}
Let $f_j^{(1)} \sim GP(0,k_1(\vec{x},\vec{x}'))$
be a gaussian process for the $1$st fidelity i.e. $m_j = 1$, where $k_1: \mathcal{R}^{d} \times\mathcal{R}^{d} \rightarrow \mathcal{R}$ is a suitable kernel.
The output for successively fidelities $m_j = 2, \ldots, M_j$ is recursively defined as
\begin{align}
 f_j^{(m_j)}(\vec{x}) & = f_j^{(m_j-1)}(\vec{x}) + f_{j_e}^{(m_j-1)}(\vec{x}), 
 \label{eq:fidelity-diff}
\end{align}
where, $f_{j_e}^{(m_j-1)} \sim GP(0, k_e(\vec{x},\vec{x}^\prime))$
with $k_e: \mathcal{R}^d \times \mathcal{R}^d \rightarrow \mathcal{R}$.
It is assumed that $f_{j_e}^{(m_j-1)}$ is conditionally independent from all fidelities lower than $m_j$. As a result, {\em the kernel for a pair of points evaluated at the same fidelity} becomes:
\begin{align}
 k_{m_j}(\vec{x},\vec{x}^\prime) \equiv k_1(\vec{x},\vec{x}^\prime) + (m_j-1) k_e(\vec{x},\vec{x}^\prime)
 \label{eq:mfgpr-kernel}
\end{align}
and as a result, the output for $m_j$th fidelity is also modeled as a gaussian process:
\begin{align*}
 f_j^{(m_j)} \sim GP(0, k_{m_j}(\vec{x},\vec{x}^\prime)).
\end{align*}

{\em The kernel function for a pair of inputs evaluated at different fidelities $m_j$ and $m_j'$ is given as}:
\begin{align*}
 k((\vec{x},m_j),(\vec{x}',m_j')) 
 & =
 \mathrm{cov}\left(
 f_j^{(m_j)}(\vec{x}),
 f_j^{(m_j')}(\vec{x}')
 \right)
 \\
 & =
 k_{m_j}(\vec{x},\vec{x}'),
\end{align*}
where $m_j \leq m_j'$ and $\mathrm{cov}$ represents covariance.
Using a kernel matrix $\*K \in \mathcal{R}^{n \times n}$ in which the $p,q$ element is defined by 
$k((\vec{x},m_j^p),(\vec{x}',m_j^q))$, 
all fidelities $f_j^{(1)}, \ldots, f_j^{(M_j)}$ can be integrated into one common gaussian process by which predictive mean and variance are obtained as 
\begin{align}
 \mu^{(m_j)}(\vec{x}) 
 & = 
 \*K + \sigma_{\rm noise}^2 \*I^{-1}
 \vec{Y}, 
 \label{eq:MF-GPR-mean}
 \\
 \sigma^{2^{(m_j)}}(\vec{x})  
 & = 
 k((\vec{x},m_j),(\vec{x},m_j))  
 \nonumber \\
 & - \*k^{(m_j)}_n(\vec{x})^\top 
 \*K + \sigma_{\rm noise}^2 \*I^{-1}
 \*k^{(m_j)}_n(\vec{x}),
 \label{eq:MF-GPR-var}
\end{align}
where 
$\vec{Y} = (y_1^{(m_{j_1})}(\vec{x}_1), \ldots, y_n^{(m_{j_n})}(\vec{x}_n))^\top$
and
$k^{(m_j)}_n(\vec{x}) $\\
$ \equiv (k((\vec{x},m_j),(\vec{x}_1,{m_j}_1))
 , \ldots, k((\vec{x},m_j),(\vec{x}_n,{m_j}_n)))^\top$.

We also define  $\sigma^{2^{(m_jm_j^\prime)}}(\vec{x})$ as the predictive covariance between $(\vec{x},m_j)$ and $(\vec{x},m_j^\prime)$, i.e., covariance for identical $\vec{x}$ at different fidelities:

\begin{align}
 \begin{split}  
  \sigma^{2(m_jm_j^\prime)} & (\vec{x}) 
   = 
  k((\vec{x},m_j),(\vec{x},m_j^\prime))  
  \\
  & - 
 \*k^{(m_j)}_n(\vec{x})^\top 
 \*K + \sigma_{\rm noise}^2 \*I^{-1}
 \*k^{(m_j^\prime)}_n(\vec{x}).
 \end{split}
 \label{eq:MF-GPR-cov}
\end{align}

\subsection{Multi-fidelity output space entropy based acquisition function} 

We describe our proposed acquisition function for multi-fidelity multi-objective setting in this section. We leverage the information-theoretic principle of output space information gain to develop an efficient and robust acquisition function. The proposed method is applicable for the general case, where at each iteration the objective functions can be evaluated at different fidelities. 

The key idea behind the proposed acquisition function is to find the pair $\{\vec{x}, \vec{m}\}$ that maximizes the information gain about the {\bf Pareto front of the highest fidelities (denoted by $\mathcal{Y}^*$)} per unit cost, where $\{\vec{x}, \vec{m}\}$ represents a candidate input $\vec{x}$ evaluated at a vector of fidelities $\vec{m} = [m_1, m_2, \cdots, m_K]$. 

\begin{figure}[t]
    \centering
    \includegraphics[width=1\columnwidth]{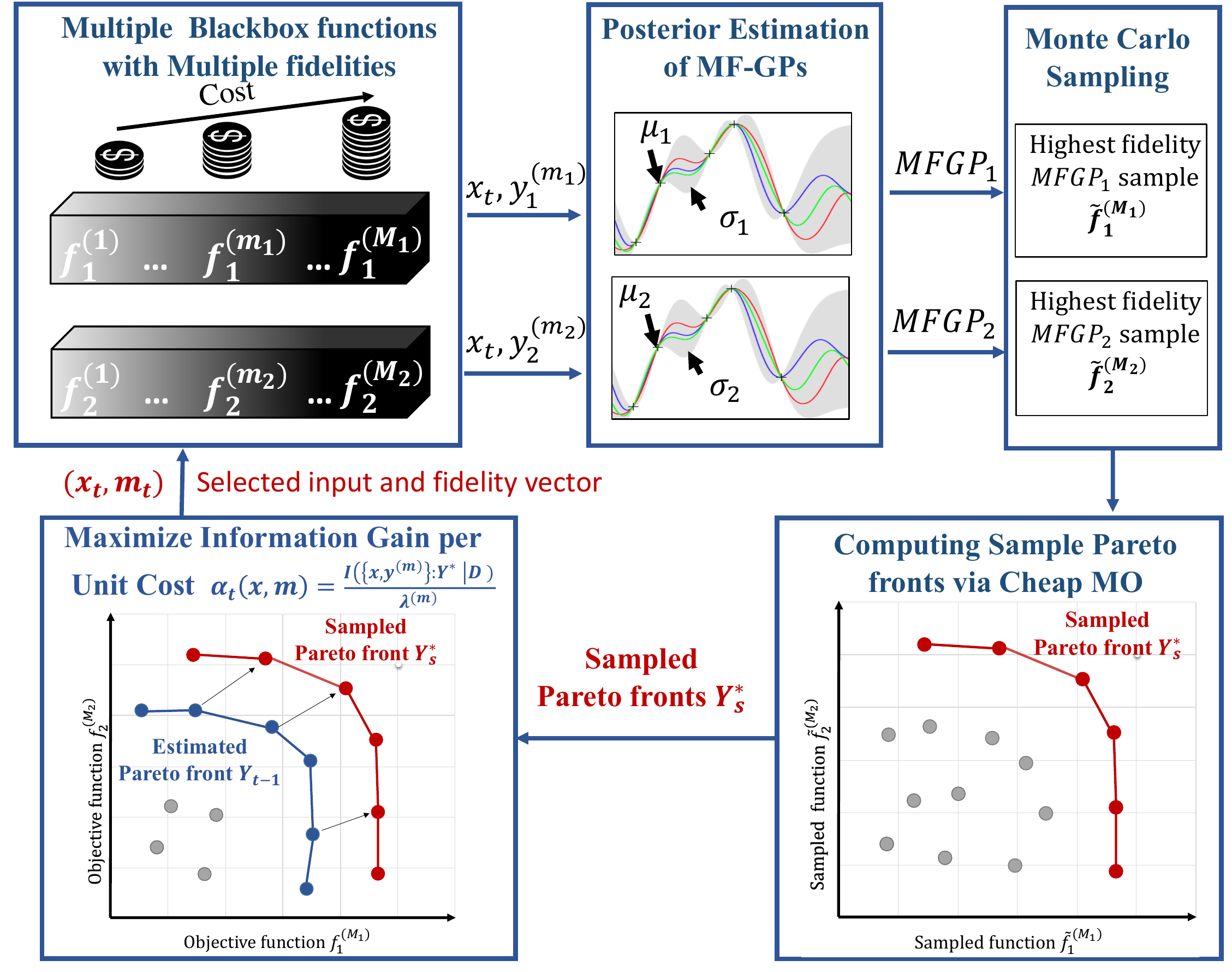}
    \caption{Overview of the MF-OSEMO algorithm for two objective functions ($k$=2). We build multi-fidelity statistical models $\mathcal{MFGP}_1$, $\mathcal{MFGP}_2$ for the two objective functions $f_1(x) $ and $f_2(x)$ with $M_1$ and $M_2$ fildelities respectively. First, we sample highest fidelity functions from the statistical models. We compute sample pareto fronts by solving a cheap MO problem over the sampled functions. Second, we select the best candidate input $x_t$ and fidelity vector $m_t=(m_1,m_2)$ that maximizes the information gain per unit cost . Finally, we evaluate the functions for $x_t$ at fidelities $m_t$ to get $(y_1^{(m_1)}, y_2^{(m_2)})$ and update the statistical models using the new training example.}
    \label{fig:algo}
\end{figure}

This idea  can be expressed mathematically 
as given below:
\begin{align}
        \alpha(\vec{x},\vec{m}) &= I(\{\vec{x}, \vec{y}^{(\vec{m})}\}, \mathcal{Y}^* \mid D) / \lambda^{(\vec{m})} \label{af:def}
\end{align}
where $\lambda^{(\vec{m})}$ is the total {\em normalized} cost of evaluating the objective functions at  $\vec{m}$ and $D$ is the data collected so far.
Figure \ref{fig:algo} provides an overview of the  MF-OSEMO algorithm.
The information gain in equation \ref{af:def} is defined as the expected reduction in entropy $H(.)$ of the posterior distribution $P(\mathcal{Y}^* \mid D)$ as a result of evaluating $\vec{x}$ at fidelity vector $\vec{m}$:
\begin{align}
    &I(\{\vec{x}, \vec{y}^{(\vec{m})}\}, \mathcal{Y}^{*} \mid D) \nonumber\\
    &= H(\mathcal{Y}^{*} \mid D) - \mathbb{E}_{y^{(\vec{m})}} [H(\mathcal{Y}^{*} \mid D \cup \{\vec{x}, \vec{y}^{(\vec{m})}\})] \label{eqn_ig_def} \\
    &= H(\vec{y}^{(\vec{m})} \mid D, \vec{x}) - \mathbb{E}_{\mathcal{Y}^{*}} [H(\vec{y}^{(\vec{m})} \mid D,   \vec{x}, \mathcal{Y}^{*})] \numberthis \label{eqn_symmetric_ig}
\end{align}
Equation \ref{eqn_symmetric_ig} follows from equation \ref{eqn_ig_def} as a result of the symmetric property of information gain. The first term in the  r.h.s of equation \ref{eqn_symmetric_ig} is the entropy of a factorizable K-dimensional gaussian distribution $P(\vec{y}^{(\vec{m})}\mid D, \vec{x}$)) which can be computed in closed form as shown below:
\begin{align}
H(\vec{y}^{(\vec{m})} \mid D, \vec{x}) = \frac{K(1+\ln(2\pi))}{2} +  \sum_{j = 1}^K  \ln (\sigma_j^{(m_j)}(\vec{x})) \label{eqn_unconditioned_entropy}
\end{align}
where $\sigma_j^{{(m_j)}}(\vec{x})$ is the predictive variance of $j^{th}$ surrogate model $GP_j$ at input $\vec{x}$ and fidelity $m_j$. The second term in the r.h.s of equation \ref{eqn_symmetric_ig} is an expectation over the Pareto front of the highest fidelities $\mathcal{Y}^{*}$. We can approximately compute this term via Monte-Carlo sampling as shown below: 

\begin{align}
    \mathbb{E}_{\mathcal{Y}^{*}} [H(\vec{y}^{(\vec{m})} \mid D,   \vec{x}, \mathcal{Y}^{*})] \simeq \frac{1}{S} \sum_{s = 1}^S [H(\vec{y}^{(\vec{m})} \mid D,   \vec{x}, \mathcal{Y}^{*}_s)] \label{eqn_summation}
\end{align}
where $S$ is the number of samples and $\mathcal{Y}^{*}_s$ denote a sample Pareto front obtained over the highest fidelity functions sample  from $K$ surrogate models. 
The main advantages of our acquisition function are: cost efficiency, computational-efficiency, and robustness to the number of samples. Our experiments demonstrate these advantages over state-of-the-art single fidelity AFs for multi-objective optimization. 

There are two key algorithmic steps to compute Equation \ref{eqn_summation}: 1) Computing Pareto front samples $\mathcal{Y}^{*}_s$; and 2) Computing the entropy with respect to a given Pareto front sample $\mathcal{Y}^{*}_s$. We provide solutions for these two steps below.

\vspace{0.8ex}

\hspace{2.0ex} {\bf 1) Computing Pareto front samples via cheap multi-objective optimization.} To compute a Pareto front sample $\mathcal{Y}^{*}_s$, we first sample highest fidelity functions from the posterior MF-GP models via random fourier features \cite{PES,random_fourier_features} and then solve a cheap multi-objective optimization over the $K$ sampled high fidelity functions. It is important to note that we are sampling only the {\bf highest fidelity function} from each MF-GP surrogate model.

\hspace{3.5ex} {\em Sampling functions from  the posterior of MF-GP model.} Similar to prior work \cite{PES,PESMO,MES}, we employ random fourier features based sampling procedure. We approximate each GP prior of the highest fidelity as $\Tilde{f}^{(M)} = \phi(\vec{x})^T \theta$, where $\theta \sim N(0, \vec{I})$. The key idea behind random fourier features is to construct each function sample $\Tilde{f}^{(M)}(\vec{x})$ as a finitely parametrized approximation: $\phi(\vec{x})^T \theta$, where $\theta$ is sampled from its corresponding posterior distribution conditioned on the data $D$ obtained from past function evaluations: $\theta | D \sim N(\vec{A^{-1}\Phi^Ty}_n, \sigma^2\vec{A^{-1}})$, where $\vec{A} = \vec{\Phi^T\Phi} + \sigma^2 \vec{I}$ and $\Phi^T = [\phi(\vec{x}_1),\cdots,\phi(\vec{x}_{t-1})]$.

\hspace{3.5ex}{\em Cheap MO solver.} We sample $\Tilde{f}^{(M_i)}_i$ from each surrogate model $\mathcal{MF-GP}_i$ as described above. A {\em cheap} multi-objective optimization problem over the $K$ sampled functions $\Tilde{f}^{(M_1)}_1,\Tilde{f}^{(M_2)}_2,\cdots,\Tilde{f}^{(M_K)}_K$ is solved to compute the sample Pareto front $\mathcal{Y}^{*}_s$. This cheap multi-objective optimization also allows us to capture the interactions between different objectives. We employ the popular NSGA-II algorithm \cite{deb2002nsga} to solve the MO problem with cheap objective functions noting that any other algorithm can be used. 

\vspace{0.8ex}

\hspace{2.0ex}{\bf 2) Entropy computation with a sample Pareto front.}
Let $\mathcal{Y}^{*}_s = \{\vec{z}^1, \cdots, \vec{z}^l \}$ be the sample Pareto front,  where $l$ is the size of the Pareto front and each $\vec{z}^i = \{z_1^i,\cdots,z_K^i\}$ is a $K$-vector evaluated at the $K$ sampled high fidelity functions. The following inequality holds for each component $y^{(m_j)}_j$ of the $K$-vector $\vec{y}^{\vec{(m)}} = \{y^{(m_1)}_1, \cdots, y^{(m_k)}_K\}$ in the entropy term $H(\vec{y}^{\vec{(m)}} \mid D,   \vec{x}, \mathcal{Y}^{*}_s)$:
\begin{align}
 y^{(m_j)}_j &\leq y_{j_s}^{*} \quad \forall j \in \{1,\cdots,K\} \label{inequality}
\end{align}
where $y_{j_s}^{*} = \max \{z^1_j, \cdots z^l_j \}$. The inequality essentially says that the $j^{th}$ component of $\vec{y}^{\vec{m}}$ (i.e., $y^{m_j}_j$) is upper-bounded by a value obtained by taking the maximum of $j^{th}$ components of all $l$ vectors $\{\vec{z}^1, \cdots, \vec{z}^l \}$ in the Pareto front $\mathcal{Y}^{*}_s$. The proof of \ref{inequality} can be divided into two cases:

\vspace{0.3ex}

{\bf Case I.} If $y_j$ is evaluated at its highest fidelity (i.e $m_j=M_j$), inequality \ref{inequality} can be proven by a contradiction argument. Suppose there exists some component $y^{(M_j)}_j$ of $\vec{y}^{(\vec{M})}$ such that $ y^{(M_j)}_j > y_{j_s}^{*}$. However, by definition, $\vec{y}^{(\vec{M})}$ is a non-dominated point because no point dominates it in the $j$th dimension. This results in $\vec{y}^{(\vec{M})} \in \mathcal{Y}^*_s$ which is a contradiction. Therefore, our hypothesis that $ y^{(M_j)}_j> y_{j_s}^{*}$ is incorrect and inequality \ref{inequality} holds.

\vspace{0.3ex}

{\bf Case II.} If $y_j$ is evaluated at one of its lower fidelities (i.e, $m_j \neq M_j$), the proof follows from the assumption that the value of lower fidelity of a objective is usually smaller than the corresponding higher fidelity, i.e., $y^{(m_j)}_j \leq y^{(M_j)}_j \leq y_{j_s}^{*}$. This is especially true for most real-world experiments. For example, in optimizing a neural network's accuracy with respect to its hyperparameters, a commonly employed fidelity is the number of data samples used for training. It is reasonable to believe that the accuracy is always higher for the higher fidelity (more data samples to train on) when compared to a lower fidelity (less data samples to train on).  

By combining the inequality \ref{inequality} and the fact that each function is modeled as an independent MF-GP, a common property of entropy measure allows us to decompose the entropy of a set of independent variables into a sum over entropies of individual variables \cite{information_theory}:
\begin{align}
H(\vec{y}^{\vec{(m)}} \mid D,   \vec{x}, \mathcal{Y}^{*}_s) \simeq \sum_{j=1}^K H(y^{(m_j)}_j|D, \vec{x},y_{j_s}^{*}) \label{eqn_sep_ineq}
\end{align}

The computation of \ref{eqn_sep_ineq} requires the computation of the entropy of $p(y^{(m_j)}_j|D, \vec{x},y_{j_s}^{*})$. This is a conditional distribution that depends on the value of $m_j$ and can be expressed as $H(y^{(m_j)}_j|D, \vec{x},y^{(m_j)}_j\leq y_{j_s}^{*})$. This entropy is dealt with in two cases:

\textbf{First, for $\mathbf{m_j=M_j}$,} the density function of this probability is a truncated Gaussian distribution and its entropy can be expressed as \cite{entropy_handbook}:
\begin{align*}
&H(y^{(M_j)}_j|D, \vec{x},y^{(M_j)}_j \leq y_{j_s}^{*})=\frac{(1 + \ln(2\pi))}{2}  + \\
&\ln(\sigma^{(M_j)}_j(\vec{x}))  + \ln \Phi(\gamma_s^{(M_j)}(\vec{x}))
-\frac{\gamma_s^{(M_j)}(\vec{x}) \phi(\gamma_s^{(M_j)}(\vec{x}))}{2\Phi(\gamma_s^{(M_j)}(\vec{x}))} \numberthis \label{Conditional_high_fidel}
\end{align*}
where $\gamma_s^{(M_j)}(\vec{x}) = \frac{y_{j_s}^{*} - \mu_j^{(M_j)}(\vec{x})}{\sigma_j^{(M_j)}(\vec{x})}$, and $\phi$ and $\Phi$ are the p.d.f and c.d.f of a standard normal distribution respectively.

\textbf{Second, for $\mathbf{m_j \neq M_j}$}, the density function of $p(y^{(m_j)}_j|D, \vec{x},y_{j_s}^{*})$ can be computed using two different approximations as described below:

\vspace{0.8ex}

\textbf{Approximation 1 (MF-OSEMO-TG): }
As a consequence of {\bf Case II}, which states that $y^{(m_j)}_j\leq y_{j_s}^{*}$ also holds for all lower fidelities, the entropy of $p(y^{(m_j)}_j|D, \vec{x},y_{j_s}^{*})$ can also be approximated by the entropy of a truncated gaussian distribution and expressed as follow:
\begin{align*}
&H(y^{(m_j)}_j|D, \vec{x},y^{(m_j)}_j\leq y_{j_s}^{*})=\frac{(1 + \ln(2\pi))}{2}+    \\
&\ln(\sigma^{(m_j)}_j(\vec{x})) +  \ln \Phi(\gamma_s^{(m_j)}(\vec{x}))- \frac{\gamma_s^{(m_j)}(\vec{x}) \phi(\gamma_s^{(m_j)}(\vec{x}))}{2\Phi(\gamma_s^{(m_j)}(\vec{x}))} \numberthis \label{TGappriximation}
\end{align*}
where $\gamma_s^{(m_j)}(\vec{x}) = \frac{y_{j_s}^{*} - \mu_j^{(m_j)}(\vec{x})}{\sigma_j^{(m_j)}(\vec{x})}$.\\

\vspace{0.5ex}

\textbf{Approximation 2 (MF-OSEMO-NI): }
Although equation \ref{TGappriximation} is sufficient for computing the entropy for $m_j \neq M_j$, it can be improved by conditioning on a tighter inequality $y^{(M_j)}_j\leq y_{j_s}^{*}$ as compared to the general one, i.e., $y^{(m_j)}_j\leq y_{j_s}^{*}$.
As we show below, this improvement comes at the expense of not obtaining a final closed-form expression, but it can be efficiently computed via numerical integration. We use the derivation of the entropy based on numerical integration, proposed in \cite{takeno2019multi}.\\
Now, for calculating $H(y^{(m_j)}_j|D, \vec{x},y^{(m_j)}_j\leq y_{j_s}^{*})$ by replacing  $p(y^{(m_j)}_j|D, \vec{x},y^{(m_j)}_j \leq y_{j_s}^{*})$ with $p(y^{(m_j)}_j|D, \vec{x},y^{(M_j)}_j\leq y_{j_s}^{*})$ and using Bayes’ theorem, we have:
\begin{align}
    &p(y^{(m_j)}_j|D, \vec{x},y^{(M_j)}_j\leq y_{j_s}^{*}) \nonumber\\
    &=\frac{ p(y^{(M_j)}_j\leq y_{j_s}^{*} | y^{(m_j)}_j,D,\vec{x}) p(y^{(m_j)}_j,D,\vec{x})}{p(y^{(M_j)}_j\leq y_{j_s}^{*}|D,\vec{x})}\label{totalprob}
\end{align}
Both the densities, $p(y^{(M_j)}_j\leq y_{j_s}^{*}|D,\vec{x})$ and $p(y^{(m_j)}_j,D,\vec{x})$ can be obtained from the predictive distribution of MF-GP model and is given as follows: 
\begin{align}
  & p(y^{(m_j)}_j,D,\vec{x})=\frac{\phi(\gamma_j^{(m_j)}(\vec{x}))}{\sigma_j^{(m_j)}} \label{prob1}\\
 &  p(y^{(M_j)}_j\leq y_{j_s}^{*}|D,\vec{x})=\Phi(\gamma_s^{(M_j)}(\vec{x}))) \label{prob2}
\end{align}
where $\gamma_j^{(m_j)}(\vec{x}) = \frac{y_j^{(m_j)} - \mu_j^{(m_j)}(\vec{x})}{\sigma_j^{(m_j)}(\vec{x})}$.

Since MF-GP represents all fidelities as one unified Gaussian process, the joint marginal
distribution $p(y^{(M_j)}_j, y^{(m_j)}_j|D,\vec{x})$ can be immediately obtained from the posterior distribution of the corresponding surrogate model $\mathcal{GP}_j$ as given below:
\begin{align}
    p(y^{(M_j)}_j| y^{(m_j)}_j,\vec{x},D) \sim \mathcal{N}(\mu_j(\vec{x}),s_j^2(\vec{x}))
    \label{eqn:marginal_gaussian}
\end{align}
where $\mu_j(\vec{x})=\frac{\sigma_j^{2^{(m_jM_j)}}(\vec{x})(y_j^{(m_j)}-\mu_j^{m_j}(\vec{x}))}{\sigma_j^{2^{(m_j)}}(\vec{x})}$ \\
and $s_j^2(\vec{x})=\sigma_j^{2^{(M_j)}}(\vec{x}) - \frac{(\sigma_j^{2^{(m_jM_j)}}(\vec{x}))^2}{\sigma_j^{2^{(m_j)}}(\vec{x})}$.
As a result, $p(y^{(M_j)}_j\leq y_{j_s}^{*} | y^{(m_j)}_j,D,\vec{x})$ is expressed as the cumulative distribution of the Gaussian in \ref{eqn:marginal_gaussian}:
\begin{align}
    p(y^{(M_j)}_j\leq y_{j_s}^{*} | y^{(m_j)}_j,D,\vec{x})=\Phi(\frac{y_{j_s}^{*}-\mu_j(\vec{x})}{s_j(\vec{x})})\label{prob3}
\end{align}
By substituting \ref{prob1}, \ref{prob2}, and \ref{prob3} into \ref{totalprob} we get:
\begin{align}
     H(y^{(m_j)}_j|D, &\vec{x},y^{(M_j)}_j\leq y_{j_s}^{*})= \nonumber \\
     &- \int \Psi(y^{(m_j)}_j) \log(\Psi(y^{(m_j)}_j))dy^{(m_j)}_j \label{NIappriximation}
\end{align}
With $\Psi(y^{(m_j)}_j)= \Phi(\frac{y_{j_s}^{*}-\mu_j(\vec{x})}{s_j(\vec{x})}) \frac{\Phi(\gamma_s^{(M_j)}(\vec{x})))\phi(\gamma_j^{(m_j)}(\vec{x}))}{\sigma_j^{(m_j)}}$\\

Since this integral is over one-dimension variable $y^{(m_j)}_j$, numerical integration can result in a tight approximation. 

\vspace{0.3ex}

A complete description of the MF-OSEMO algorithm is given in Algorithm \ref{alg:OSEMO}. The blue colored steps correspond to computation of our 
acquisition function via sampling.

\begin{algorithm*}[t]
\footnotesize
\caption{MF-OSEMO Algorithm}
\label{alg:OSEMO}
\textbf{Input}: input space $\mathfrak{X}$; $K$ blackbox objective functions where each function $f_j$ has multiple fidelities $M_j$ $\left(\{f_1^{(1)}(\vec{x}), \cdots, f_1^{(M_1)}(\vec{x})\},
\cdots,\{f_K^{(1)}(\vec{x}), \cdots, f_K^{(M_K)}(\vec{x})\}\right)$; and  total cost budget $\lambda_{Total}$
\begin{algorithmic}[1] 
\STATE Initialize multi-fidelity gaussian process surrogate models $\mathcal{GP}_1, \cdots, \mathcal{GP}_K$ by evaluating at initial points $D$
\STATE \textbf{While} {$\lambda_{t} \leq \lambda_{total}$} \textbf{do}
 \STATE \quad \textcolor{blue}{for each sample $s \in {1,\cdots,S}$:} 
 \STATE \quad \quad \textcolor{blue}{Sample highest-fidelity functions $\Tilde{f}_i^{(M_i)} \sim \mathcal{GP}_i, \quad \forall{i \in \{1,\cdots, K\}} $}
 \STATE \quad \quad \textcolor{blue}{$\mathcal{Y}_s^{*} \leftarrow$ Pareto front of {\em cheap} multi-objective optimization over $(\Tilde{f}_1^{(M_1)}, \cdots, \Tilde{f}_K^{(M_K)})$}
 \STATE \quad \textcolor{blue}{Find the next point to evaluate: select $(\vec{x}_{t},\vec{m}_t) \leftarrow \arg max_{\vec{x}\in \mathfrak{X},\vec{m}} \hspace{2 mm} \alpha_t(\vec{x},\vec{m},\mathcal{Y}^{*}) $}
 \STATE \quad Update the total cost consumed: $\lambda_t \leftarrow \lambda_t + \lambda^{\vec{(m_t)}}$
\STATE \quad Aggregate data: $\mathcal{D} \leftarrow \mathcal{D} \cup \{(\vec{x}_{t}, \vec{y}_{t}^{\vec{m}})\}$ 
\STATE \quad Update models $\mathcal{GP}_1,\cdots, \mathcal{GP}_K$ 
\STATE \quad $t \leftarrow t+1$
\STATE \textbf{end while}
\STATE \textbf{return} Pareto front and Pareto set of $f_1(x), \cdots,f_K(x)$ based on $\mathcal{D}$\\
\STATE \textbf{end}
\STATE \textbf{Procedure} {$\alpha_t(\vec{x},\vec{m},\mathcal{Y}_s^{*})$}{}
\STATE  // Computes information gain (IG) about the posterior of true Pareto front $(\mathcal{Y}^*)$ per unit cost as a result of evaluating $\vec{x}$
\STATE // IG = $H_1$ - $H_2$;\quad where $H_1$ = Entropy of $\vec{y}^{(\vec{m})}$ conditioned on $D$ and $\vec{x}$ \\
//\hspace{30mm} and $H_2$ = Expected entropy of $\vec{y}^{(\vec{m})}$ conditioned on $D$, $\vec{x}$ and $(\mathcal{Y}^*)$ 
\STATE Set $H_1 = H(\vec{y}^{(\vec{m})} \mid D, \vec{x}) = {K(1+\ln(2\pi))}/{2} +  \sum_{j = 1}^K  \ln (\sigma_j^{(m_j)}(\vec{x})) $ (entropy of K-factorizable Gaussian)
\STATE To compute $H_2  \simeq \frac{1}{S} \sum_{s = 1}^S \sum_{j=1}^K H(y^{(m_j)}_j|D, \vec{x},y_{j_s}^{*})$, initialize $H_2$ = 0
\FOR{each sample $\mathcal{Y}_s^*$}
\FOR{$j \in {1 \cdots K}$}
\STATE Set $y_{j_s}^{*}$ = maximum of $j$th component of all vectors in $\mathcal{Y}_s^*$
\STATE \textbf{If }{$m_j=M_j$} \quad // if evaluating $j$th function at highest fidelity
\STATE \quad  $H_2$ += $H(y^{(M_j)}_j|D, \vec{x},y^{(M_j)}_j \leq y_{j_s}^{*})$ (entropy of truncated Gaussian $p(y^{(M_j)}_j|D, \vec{x},\underline{y^{(M_j)}_j \leq y_{j_s}^{*}})$)
\STATE \textbf{Else if }{$m_j\neq M_j$} \quad // if evaluating $j$th function at lower fidelity
\STATE \quad \quad // two approximations are provided
\STATE \quad \quad {\bf If} approximation = TG 
\STATE \quad \quad \quad  $H_2$ += $H(y^{(m_j)}_j|D, \vec{x},y^{(m_j)}_j\leq y_{j_s}^{*})$ (entropy of truncated Gaussian $p(y^{(M_j)}_j|D, \vec{x},\underline{y^{(m_j)}_j \leq y_{j_s}^{*}})$)
\STATE \quad \quad {\bf Else If} approximation = NI 
\STATE \quad \quad \quad  $H_2$ += $H(y^{(m_j)}_j|D, \vec{x},y^{(M_j)}_j\leq y_{j_s}^{*})$ (entropy computed via numerical integration)
\ENDFOR
\ENDFOR
\STATE  Divide by number of samples: $H_2$ = $H_2/S$
\STATE \textbf{return $(H_1 - H_2)/\lambda^{(\vec{m})}$} 
\end{algorithmic}
\end{algorithm*}

\section{Experiments and Results}
In this section, we describe our experimental setup, and present results of MF-OSEMO and baseline methods.

\subsection{Experimental Setup}

\vspace{0.3ex}
\noindent {\bf Baselines.} We compare MF-OSEMO with state-of-the-art single-fidelity MO algorithms: ParEGO \cite{knowles2006parego}, PESMO \cite{PESMO}, SMSego \cite{ponweiser2008multiobjective}, EHI \cite{emmerich2008computation}, and SUR \cite{picheny2015multiobjective}. We employ the code for these methods from the BO library Spearmint\footnote{https://github.com/HIPS/Spearmint/tree/PESM}. 

\vspace{0.3ex}
\noindent {\bf Statistical models.} We use MF-GP models as described in section \ref{surrogatesection}. We employ squared exponential (SE) kernels in all our experiments. 
 The hyper-parameters are estimated after every 5 function evaluations. We initialize the MF-GP models for all functions by sampling initial points at random from a Sobol grid. We Initialise each of the lower fidelities with 5 points and the highest fidelity with only one point.
 
\vspace{0.3ex}
\noindent {\bf Synthetic benchmarks.} We construct two synthetic benchmark problems using a combination of commonly employed benchmark functions for multi-fidelity and single-objective optimization \footnote{https://www.sfu.ca/~ssurjano/optimization.html}, and two of the known general MO benchmarks \cite{habib2019multiple}. Their complete details are provided in Table \ref{tab:synth}.

\vspace{0.3ex}
\noindent {\bf Real-world benchmarks.} We consider two challenging problems that are described below. 

\vspace{0.3ex}
{\bf 1) Rocket launching simulation.} We consider the simulation study of a rocket \cite{hasbun2012classical} being launched from the Earth’s surface. 
Input variables for simulation are mass of fuel, launch height, and launch angle. 
Output objectives are the time taken to return to Earth's surface, the angular distance travelled with respect to the centre of the Earth, and the absolute difference between the launch angle and the radius at the point of launch. However, these simulations are computationally expensive and can take up to several hours. The simulator has a tolerance parameter that can be adjusted to perform multi-fidelity simulations: small tolerance means accurate simulations, but long runtime. We employ two tolerance parameter values to create two fidelities for each objective: cost of two fidelities are 0.05 minutes and 30 minutes respectively.

\vspace{0.3ex}
{\bf 2) Network-on-chip (NOC) optimization.} Designing good communication infrastructure is important to improve the quality of hardware designs. This is typically done using cycle-accurate simulators that imitate the real hardware. We consider a design space of NoC dataset consisting of 1024 implementation of a network-on-chip \cite{rodinia-benchmark}. Each configuration is defined by ten input variables ($d$=10). We optimize two objectives: latency and energy. This benchmark has two fidelities with costs 3 mins and 45 mins respectively. 
\begin{table}[h]
\centering
\resizebox{0.9\linewidth}{!}{
\begin{tabular}{llllll}  
\toprule
Name &  $k$ & $d$ &Benchmark functions & $p  $ & Costs \\
\midrule
BC  & 2   &  2  & \begin{tabular}{l}Branin \\Currin\end{tabular}    & \begin{tabular}{r}2 \\2\end{tabular}  & \begin{tabular}{l} $[1,10]$ \\ $[1,10]$\end{tabular}  \\
\midrule
SPP & 3 & 4  &  \begin{tabular}{l}Shekel \\Park 1 \\ Park 2\end{tabular}   &  \begin{tabular}{l}3 \\2 \\  2\end{tabular}  &  \begin{tabular}{l}$[0.1,1,10]$ \\$[1,10]$ \\ $[1,10]$\end{tabular} \\
\midrule
ZDT3 & 2  &  6 & Zitzler,Deb,Thiele  & $2^2$  &  $[1,10]^2$ \\
\midrule
DTLZ1 &  6 & 5 & Deb,Thiele,Laumanns,Zitzler & $3^6$    & $[0.1,1,10]^6$  \\
\bottomrule
\end{tabular}}
\caption{Details of synthetic benchmarks: Name, benchmark functions, no. of objectives $k$,  input dimension $d$, number fidelities $p$, and costs of different fidelities for each function.}
\label{tab:synth}
\end{table}
\subsection{Results and Discussion}
To evaluate the performance of MF-OSEMO, we employ a common multi-objective metric used in practice. The {\em Pareto hypervolume (PHV)} metric measures the quality of a given Pareto front \cite{zitzler1999evolutionary}. PHV is defined as the volume between a reference point and the given Pareto front (set of non-dominated points). As a function of the cost of evaluations, we report the difference between the hypervolume of the ideal Pareto front $(\mathcal{Y^{*}}) $ and hypervolume of the best reached Pareto front estimated by optimizing the posterior mean of the models at the highest fidelities \cite{PESMO}. The posterior means are optimized over a randomly generated grid of 10,000 points. We also provide the \textit{cost reduction factor}, which is the ratio between the worst cost at which MF-OSEMO converges (worst case for MF-OSEMO), and the earliest cost for which any of the single-fidelity baselines converge (best case for baseline) after running all algorithms for very large costs.
We run all experiments 10 times. The mean and variance of the $PHV$ metrics across different runs are reported as a function of the total cost consumed. Since in all our experiments, the costs of different functions are on the same scale, we plot results against the sum of the costs. 
 \begin{figure}[t!] 
    \centering
    \begin{minipage}{1\columnwidth}
    \centering
    \begin{minipage}{0.5\columnwidth}
        \centering
        \includegraphics[width=0.97\columnwidth]{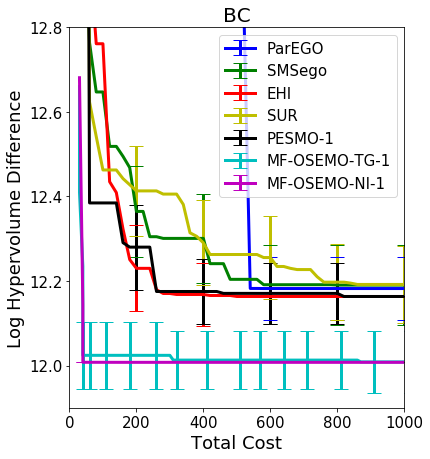} 
    \end{minipage}\hfill
    \begin{minipage}{0.5\columnwidth}
        \centering
        \includegraphics[width=0.97\columnwidth]{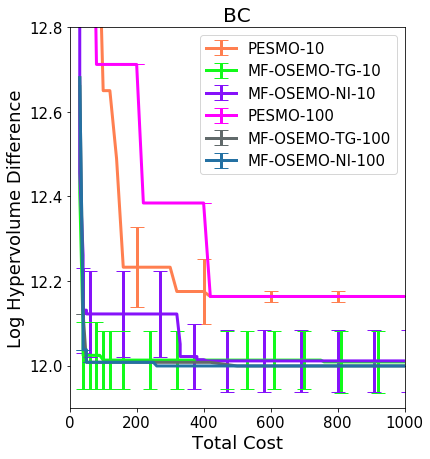} 
    \end{minipage}
    \end{minipage}
    \begin{minipage}{1\columnwidth}
            \centering
    \begin{minipage}{0.5\columnwidth}
        \centering
        \includegraphics[width=0.97\columnwidth]{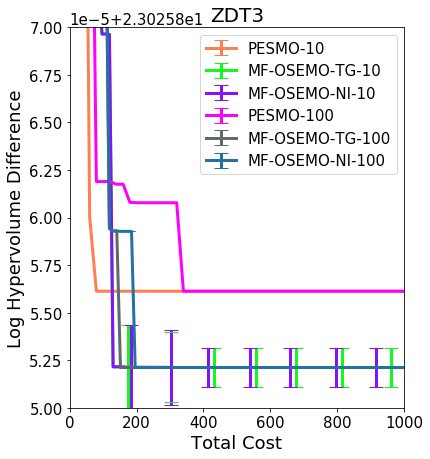} 
    \end{minipage}\hfill
    \begin{minipage}{0.5\columnwidth}
        \centering
        \includegraphics[width=0.97\columnwidth]{ZDT3_log_mean_var_phv_int_new.png} 
    \end{minipage}
    \end{minipage}

       \begin{minipage}{1\columnwidth}
    \centering
    \begin{minipage}{0.5\columnwidth}
        \centering
        \includegraphics[width=0.97\columnwidth]{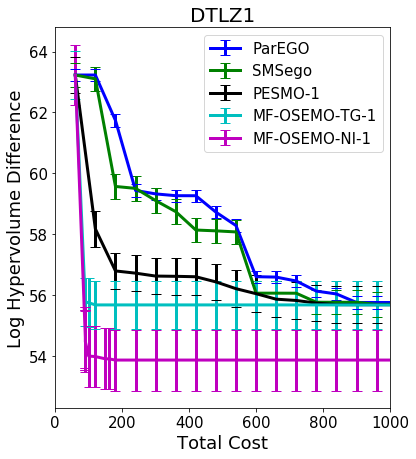} 
    \end{minipage}\hfill
    \begin{minipage}{0.5\columnwidth}
        \centering
        \includegraphics[width=0.97\columnwidth]{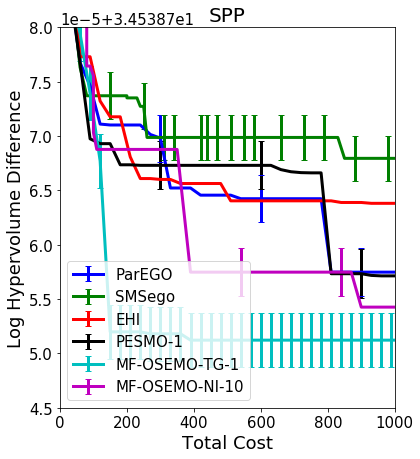} 
    \end{minipage}
    \end{minipage}
\caption{Results of MF-OSEMO and single-fidelity multi-objective BO algorithms on synthetic benchmarks. The log of the hypervolume difference is shown with varying cost.}
\label{Synthetic}
\end{figure}

\begin{figure}[t] 
    \centering
    \begin{minipage}{1\columnwidth}
    \centering
    \begin{minipage}{0.5\columnwidth}
        \centering
        \includegraphics[width=0.97\columnwidth]{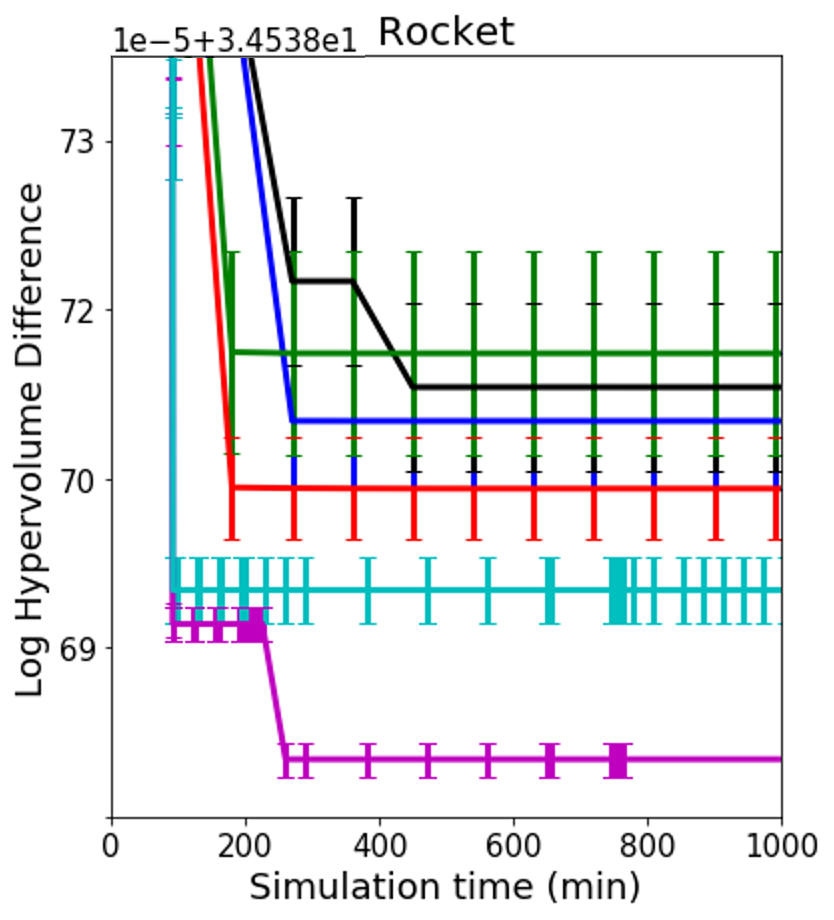} 
    \end{minipage}\hfill
    \begin{minipage}{0.5\columnwidth}
        \centering
        \includegraphics[width=0.97\columnwidth]{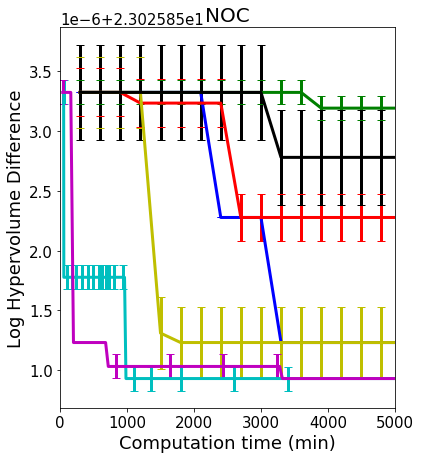} 
    \end{minipage}
    \end{minipage}
\caption{Results of MF-OSEMO and single-fidelity multi-objective BO algorithms on real-world problems. The log of the hypervolume difference is shown with varying cost.}
\label{realworld}
\end{figure}

\vspace{0.3ex}
\noindent {\bf MF-OSEMO vs. State-of-the-art.}  We compare the performance of MF-OSEMO-TG and MF-OSEMO-NI with single-fidelity MO methods. Figure \ref{Synthetic} and Figure \ref{realworld} show the results of all multi-objective BO algorithms including MF-OSEMO for synthetic and real-world benchmarks respectively. We observe that: 1) MF-OSEMO consistently performs better than all baselines. Both the variants of MF-OSEMO converge at a much lower cost. 2) Rates of convergence of MF-OSEMO-TG and MF-OSEMO-NI are slightly different. However, in all cases, MF-OSEMO performs better than baseline methods. We notice that in few cases (e.g., both real-world benchmarks), MF-OSEMO-TG converges earlier than MF-OSEMO-NI. This demonstrates that even with loose approximation, using the MF-OSEMO-TG can provide consistently competitive results using less computation time.
 \begin{table}[ht!]
\centering
\resizebox{1\linewidth}{!}{
\begin{tabular}{lllllll}  
\toprule
Name & BC  & SPP  &ZDT3 & DTLZ1 & Rocket & NOC \\
\midrule
$\lambda$ & 4.2  & 190  &380 & 100 & 250 & 1200 \\
\midrule
$\lambda_B$ & 2000  & 1950  &2000 & 800 & 4000 & 10000 \\
\midrule
$\Lambda$ & 99.79\%  & 90.25\%  &81\% & 87.5\% & 93.75\% & 88\% \\
\bottomrule
\end{tabular}}
\caption{Convergence costs for MF-OSEMO and baselines, and cost reduction factor achieved by MF-OSEMO: {\em worst} convergence cost for MF-OSEMO $\lambda$, {\em best} convergence cost from all baselines methods $\lambda_B$, and cost reduction factor $\Lambda$.}
\label{tab:costreduction}
\end{table} 

\vspace{0.3ex}
\noindent {\bf Cost reduction factor.} Some of the baselines will eventually converge if they are run for a much larger cost. In table \ref{tab:costreduction}, we provide the cost reduction factor to show the percentage of cost-gain achieved by using MF-OSEMO when compared to single-fidelity baselines. Although the metric gives advantage to  baselines, the results in the table show a consistently high gain ranging from $81\%$ to $99.8\%$. 

\vspace{0.3ex}
\noindent {\bf Robustness of MF-OSEMO.} We evaluate the performance of MF-OSEMO and PESMO with different number of Monte-Carlo samples (MCS). We provide results for two synthetic benchmarks BC and ZDT3 in figure \ref{Synthetic} with 1, 10, and 100 MCS for PESMO, MF-OSEMO-TG, and MF-OSEMO-NI. For clarity of figures, we provided those results in two diffrent figures side by side. We notice that the convergence rate of PESMO is dramatically affected by the number of Monte-Carlo samples: 100 samples lead to better results than 10 and 1. However, MF-OSEMO-TG and MF-OSEMO-NI maintain a better performance consistently even with a single sample. These results strongly demonstrate that our proposed method is much more robust to the number of MCS.

\section{Summary and Future Work}
We introduced a novel and principled approach referred as MF-OSEMO to solve multi-fidelity multi-objective Bayesian optimization problems. The key idea is to employ an output space entropy based acquisition function to efficiently select inputs and fidelity vectors for evaluation. Our experimental results on both synthetic and real-world benchmarks showed that MF-OSEMO yields consistently better results than state-of-the-art single-fidelity methods.
Immediate future work will be to apply MF-OSEMO to novel real-world applications. 

\vspace{1.0ex}

\noindent {\bf Acknowledgements.} This research is supported by National Science Foundation grants IIS-1845922 and OAC-1910213.


\end{document}